\documentclass[10pt,twocolumn,letterpaper]{article}

\usepackage{cvpr} % uncomment this for the final submission
\usepackage{times}
\usepackage{epsfig}
\usepackage{graphicx}
\usepackage{amsmath}
\usepackage{amssymb}
\usepackage{amsfonts}
\usepackage{algorithmic}
\usepackage{textcomp}
\usepackage{xcolor}
\usepackage{booktabs}
\usepackage{multirow}
\usepackage{subcaption}
\usepackage{comment}
\usepackage{caption}% for more cell option / big table lines
\usepackage{makecell}% multi colums for tables
\usepackage{adjustbox}
\usepackage[backend=bibtex, bibstyle=ieee, citestyle=numeric-comp, url=false]{biblatex}
\usepackage[breaklinks=true,letterpaper=true,hidelinks,bookmarks=false]{hyperref}

\begin{document}

%%%%%%%%% TITLE
\title{A Responsible Face Recognition Approach for Small and Mid-Scale Systems Through Personalized Neural Networks}

%\and

% for ArXiv submission
\author{Sebastian Groß, Stefan Heindorf, Philipp Terhörst \\
Paderborn University, Germany \\
%{\tt\small Corresponding Author: rouqaiah.al.refai@uni-paderborn.de}
}

\maketitle
\thispagestyle{empty}

%%%%%%%%% ABSTRACT
\begin{abstract}
   Traditional face recognition systems rely on extracting fixed face representations, known as templates, to store and verify identities. These representations are typically generated by neural networks that often lack explainability and raise concerns regarding fairness and privacy. In this work, we propose a novel model-template (MOTE) approach that replaces vector-based face templates with small personalized neural networks. This design enables more responsible face recognition for small and medium-scale systems. During enrollment, MOTE creates a dedicated binary classifier for each identity, trained to determine whether an input face matches the enrolled identity. Each classifier is trained using only a single reference sample, along with synthetically balanced samples to allow adjusting fairness at the level of a single individual during enrollment. Extensive experiments across multiple datasets and recognition systems demonstrate substantial improvements in fairness and particularly in privacy. Although the method increases inference time and storage requirements, it presents a strong solution for small- and mid-scale applications where fairness and privacy are critical.
\end{abstract}

%\begin{IEEEkeywords}
    %biometrics, face recognition, privacy, fairness, explainability, responsible AI, %personalized neural networks
%\end{IEEEkeywords}

%%%%%%%%% BODY TEXT
\section{Introduction}
Face recognition technology has emerged as a cornerstone of modern biometric systems, fundamentally transforming security, authentication, and human-computer interactions~\cite{liHandbookFaceRecognition2024}. Recent advances in deep learning, particularly through architectures like Convolutional Neural Networks (CNNs)~\cite{schroffFaceNetUnifiedEmbedding2015, dengArcFaceAdditiveAngular2022} and
Transformers~\cite{zhongFaceTransformerRecognition2021, sunPartbasedFaceRecognition2022},
have pushed face recognition accuracy to unprecedented levels, rivaling and exceeding human capabilities~\cite{sunPartbasedFaceRecognition2022}.
Despite recent advances, face recognition systems continue to face critical challenges related to responsibility, particularly in terms of privacy, fairness, and explainability.

In terms of privacy, deep learning–based face templates often contain more than just identity-related information \cite{DBLP:journals/tbbis/TerhorstFDKK21, DBLP:conf/icb/TerhorstFDKK20}. Attributes such as gender, age, ethnicity, and even health status can be inferred from these templates \cite{dantchevaWhatElseDoes2016, osorio-roigAttackFacialSoftBiometric2022}. Since these representations are typically intended for recognition purposes only, the presence of such additional personal information poses significant privacy risks \cite{DBLP:journals/tifs/MedenRTDKSRPS21}. Unauthorized access to this data can result in unfair or discriminatory treatment, undermining individual rights and trust in the technology.

The presence of this information might also lead to fairness issues \cite{FRVT2019, terhorstComprehensiveStudyFace2022, howardEvaluatingProposedFairness2023}. Most biometric systems rely on representation learning strategies that are optimized for overall recognition performance. However, these strategies are highly sensitive to the distribution of the training data properties. As a result, the system’s accuracy can vary across different population groups, leading to biased outcomes \cite{drozdowskiDemographicBiasBiometrics2020}. This becomes particularly problematic in critical applications such as law enforcement or forensic investigations, where such disparities can have serious consequences.

For these critical decision-making processes, explainability remains a substantial challenge as well. Most systems provide only a match score or decision, without insight into how that result was reached. This lack of transparency makes it difficult to interpret or contest outcomes, especially in high-stakes settings. Clear, understandable reasoning is essential to support trust and responsibility in these technologies \cite{wangDeepFaceRecognition2021, DBLP:conf/cvpr/BoraTVRR24, DBLP:conf/eccv/WillifordMB20, huberEfficientExplainableFace2024, DBLP:journals/tbbis/TerhorstHDKRK23}.

%\begin{enumerate}
%    \item \textbf{Privacy concerns}: Current systems struggle with privacy preservation, particularly regarding the protection of
%    sensitive demographic attributes encoded within facial features \cite{dantchevaWhatElseDoes2016}. Traditional approaches that map
%    facial images into a unified embedding space inadvertently preserve information about gender, age, and ethnicity, making
%    these systems vulnerable to unauthorized inference attacks \cite{osorio-roigAttackFacialSoftBiometric2022}.
        
%    \item \textbf{Algorithmic bias}: These systems often exhibit varying levels of accuracy across different demographic
%    groups~\cite{buolamwiniGenderShadesIntersectional2018}. Research has revealed persistent disparities in recognition performance based on
%    gender, age, and skin tone, with error rates often significantly higher for minority groups~\cite{drozdowskiDemographicBiasBiometrics2020}.
        
 %   \item \textbf{Lack of explainability}: The decision-making processes of modern face recognition systems remain largely non-transparent.
 %   As these systems increasingly influence critical decisions, their lack of explainability presents a significant challenge to building
 %   trust and ensuring accountability~\cite{wangDeepFaceRecognition2021}.
%\end{enumerate}

To address these fundamental challenges simultaneously, we propose an alternative solution in the form of personalized neural networks. Rather than training a single model to handle all identities by extracting and storing templates, our approach creates and stores model-templates (MOTE) in the form of a dedicated small-scale binary classifier for each enrolled individual. These personalized neural network classifiers can decide if the probe identity belongs to the enrolled reference identity. 
To overcome the challenge of training the model with only one positive reference sample, we introduce a pipeline to create synthetic training data that further allows to balance the training data and thus, to adjust the fairness at the level of a single individual.
With extensive experimentation across multiple datasets and face recognition systems, the results show that changing from storing vector-templates to model-templates can significantly enhance fairness and privacy-protection.
Moreover, since our approach transforms face recognition from a complex representation learning task to a set of simple binary classification tasks, well-researched approaches for explainability can be applied to get more understandable decisions.
Our reliable face recognition approach shows strong advantages in terms of privacy, fairness, and explainability at the cost of higher storage time and computation requirements, making our MOTE approach highly useful for small- and mid-scale recognition systems that prioritize responsibility factors such as fairness and privacy.

%The core innovation lies in treating face recognition as a collection of person-specific binary classification problems rather than a unified
%embedding-based matching system. For each enrolled identity, a dedicated neural network is trained that learns to distinguish that specific
%individual from all others. This architectural choice offers several key advantages over traditional approaches. By isolating each identity's
%recognition process, information is naturally compartmentalized, making it more difficult to extract demographic attributes through inference
%attacks. Since each network specializes in recognizing only one person, it needs to learn only those features most relevant for that specific
%identity discrimination task, potentially discarding sensitive soft-biometric information.

\section{Related Work}
Current face recognition systems predominantly leverage deep learning (DL) models,
primarily Convolutional Neural Networks (CNNs) and more recently Transformers, to achieve high
accuracy~\cite{wangDeepFaceRecognition2021}. The standard paradigm involves training these complex
models to map facial images onto low-dimensional, discriminative feature vectors, commonly referred
to as templates or embeddings~\cite{jain2007handbook,schroffFaceNetUnifiedEmbedding2015, dengArcFaceAdditiveAngular2022, mengMagFaceUniversalRepresentation2021}.
While highly effective in terms of recognition accuracy, this prevailing template-based DL approach presents
significant challenges concerning algorithmic fairness, data privacy, and system explainability~\cite{wangDeepFaceRecognition2021}.
The learned embeddings, designed to capture identity, often inadvertently encode sensitive demographic attributes \cite{DBLP:conf/icb/TerhorstFDKK20, DBLP:journals/tbbis/TerhorstFDKK21}
(like gender, age, or race), leading to potential biases \cite{terhorstComprehensiveStudyFace2022} in performance across different groups and raising privacy
concerns if templates are compromised or misused~\cite{dantchevaWhatElseDoes2016, DBLP:journals/tifs/MedenRTDKSRPS21}. Furthermore, the inherent
complexity and ``black-box'' nature of deep neural networks make it difficult to understand or explain why a
particular recognition decision was made.

\subsection{Privacy in Face Recognition}
    
The inadvertent encoding of demographic attributes in face embeddings has spurred research into privacy-enhancing techniques, as attackers might be able to extract privacy-sensitive information from the stored templates.
Privacy-enhancing techniques usually aim to preserve the recognition performance while making it hard to extract information by either (a) concealing attribute patterns \cite{terhorstUnsupervisedPrivacyenhancementFace2019}, (b) changing the recognition process  and the stored information \cite{terhorstUnsupervisedEnhancementSoftbiometric2020, terhorstPEMIUTrainingFreePrivacyEnhancing2020}, or (c) by directly removing the attribute patterns \cite{terhorstSuppressingGenderAge2019, Melzi_2023_WACV, DBLP:journals/pami/MoralesFVT21} or indirectly removing these patterns through disentangling identity and attribute information \cite{bortolatoLearningPrivacyenhancingFace2020, 10581923}.

%Terhörst et
%al. \cite{terhorstUnsupervisedPrivacyenhancementFace2019} introduced Cosine-Sensitive Noise (CSN) transformation to suppress soft-biometric information while
%    preserving recognition performance. Their Incremental Variable Elimination (IVE) approach \cite{terhorstSuppressingGenderAge2019} systematically
%removes components related to sensitive attributes from feature vectors.
    
%Unsupervised approaches like Negative Face Recognition \cite{terhorstUnsupervisedEnhancementSoftbiometric2020} and Minimum Information
%Units \cite{terhorstPEMIUTrainingFreePrivacyEnhancing2020} attempt to address privacy concerns without requiring predefined attribute labels. Bortolato et
%al. \cite{bortolatoLearningPrivacyenhancingFace2020} propose PFRNet to disentangle identity and attribute information through feature separation.
%Semi-adversarial approaches like PrivacyNet \cite{mirjaliliPrivacyNetSemiAdversarialNetworks2020} actively remove sensitive information during training.% process.

% key message
Despite these advances, recent work by Osorio-Roig et al. \cite{osorio-roigAttackFacialSoftBiometric2022} demonstrated that privacy-enhancement techniques remain vulnerable to sophisticated inference attacks, underscoring the need for stronger protections against them.

% demonstrating a strong need for more effective privacy protection against inference attacks.
    
\subsection{Fairness in Biometric Systems}
    
Demographic performance disparities in face recognition systems have been extensively documented. Drozdowski et al. \cite{drozdowskiDemographicBiasBiometrics2020} survey sources of bias in biometric systems, highlighting issues in data collection,
algorithm development, and evaluation protocols.
    
Mitigation strategies include dataset balancing approaches \cite{karkkainenFairFaceFaceAttribute2019}, fairness-enhancing model based approaches \cite{10744511, 10904287, 10744457}, and supervised \cite{DBLP:conf/wacv/KotwalM24} and unsupervised score normalization approaches \cite{terhorstPostcomparisonMitigationDemographic2020, DBLP:conf/wacv/Al-RefaiHBT25}.
%regularization~\cite{DBLP:conf/wacv/KotwalM24} and fair score normalization \cite{ terhorstPostcomparisonMitigationDemographic2020, DBLP:conf/wacv/Al-RefaiHBT25}, which applies demographic-specific thresholds to improve
%fairness without requiring model retraining. 
These approaches attempt to achieve equitable performance within the constraints of the universal embedding paradigm.
% key message
While strong advancements have been made, the complex nature of the large neural networks utilized has made it difficult to address fairness, especially on the individual level.

\subsection{Explainability in Face Recognition}

Understanding model decisions remains challenging, especially for face recognition systems based on embeddings and similarity metrics rather than explicit classifications.

White-box methods like xSSAB~\cite{huberEfficientExplainableFace2024} and EBP~\cite{DBLP:conf/eccv/WillifordMB20} analyze gradient flows to identify important facial regions for matching decisions, but require access to model internals. Black-box approaches~\cite{meryBlackBoxExplanationFace2022, DBLP:conf/cvpr/BoraTVRR24} generate explanations by perturbing inputs and observing output changes without architectural knowledge.
%For example, Bora et al.\cite{DBLP:conf/cvpr/BoraTVRR24} apply LIME to generate explanations in terms of superpixels, highlighting which regions contribute most to a match score.
Whereas the former methods focus on explaining single decisions, Wang et al. \cite{wangDeepFaceRecognition2021} explain what single layers of a face recognition system learn, e.g., learning different levels of abstraction from edges and textures to high-level semantic facial attributes, revealing the hierarchical structure of face recognition models.

%%% Old.
%Understanding the decision-making process of face recognition systems remains challenging, especially for universal embedding models that rely on similarity metrics rather than explicit classification. White-box methods like xSSAB \cite{huberEfficientExplainableFace2024} analyze gradient flows to identify important facial regions for matching decisions, but require access to model internals. Black-box approaches \cite{meryBlackBoxExplanationFace2022} analyze system responses to image modifications without needing architectural knowledge.

Standard explainability techniques developed for classification tasks, such as Grad-CAM~\cite{selvarajuGradCAMVisualExplanations2020} and Grad-CAM++~\cite{chattopadhyayGradCAMImprovedVisual2018}, cannot always be directly applied to embedding-based face recognition systems due to their similarity-based architecture.
% key message
The strong advancement of explainability techniques in machine learning is mostly limited to simple classifiers and not easily adaptable to more complex representation-learning tasks. 

With MOTE, we present a way to overcome this transfer problem, as well as, provide a solution for effectively achieving fairer decisions and strong privacy protection against inference attacks.

\section{Methodology}
    
\subsection{System Overview}
\begin{figure}[h]
    \centering
    \includegraphics[width=0.5\textwidth]{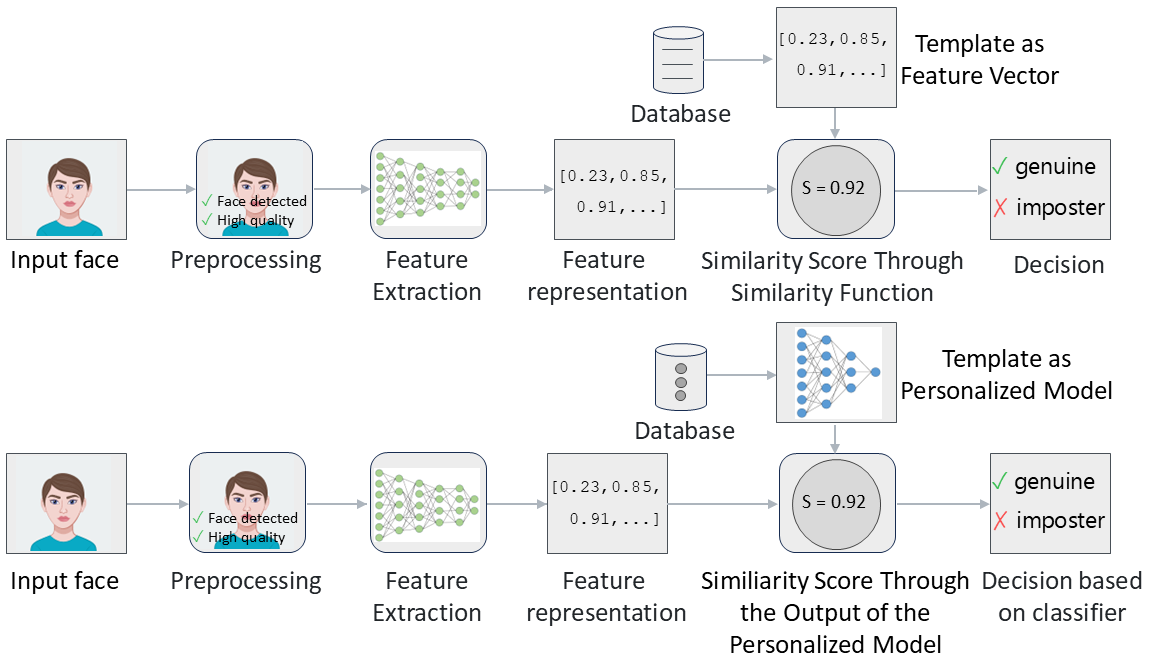}
    \caption{\textbf{Verification process of the proposed and the traditional face recognition approach}: (a) Our proposed model-template (MOTE) method
    uses a classifier-based approach where individual neural networks are trained for each identity to make genuine/imposter
    decisions directly from extracted face features; (b) Traditional similarity-based face recognition compares extracted
    templates against stored templates using a similarity function. The key difference lies in how verification decisions
    are made: MOTE leverages identity-specific models that learn to distinguish between genuine and imposter features, while
    similarity-based methods rely on fixed distance metrics between feature vectors.}
    \label{fig:workflows_comparison}
\end{figure}

In contrast to traditional face recognition systems that store and compare face templates in the form of feature vectors, our proposed approach replaces these vector-based templates with model-templates (MOTE). 
The model-template of an identity $i$ takes a vector-embedding as an input and outputs a score indicating the degree to which the input belongs to identity $i$.
Our proposed system consists of three main components: (1) a face recognition model to extract face templates, (2) a template generation procedure based on Kernel Density Estimation (KDE) to enable the training of (3) personalized verification models (binary classifiers) with only one reference sample.
%\newpage % TODO NEWPAGE EINGEFÜGT, ÜBERPRÜFEN OB DAS PASSEND IST UND ÄHNLICHES
Figure~\ref{fig:workflows_comparison} contrasts the workflow of the proposed classifier-based MOTE method with that of the existing traditional vector-based approach.

%For training our MOTE model, 

%For initial feature extraction, we utilize a face recognition feature extractor to generate face embeddings. These embeddings serve
%as templates, with each template representing a 512-dimensional feature vector. The key innovation lies in using
%these templates as input to person-specific neural networks rather than directly comparing them via similarity metrics.
    
\subsection{Template Generation using KDE}
\begin{figure}[h]
    \centering
    \includegraphics[width=\columnwidth]{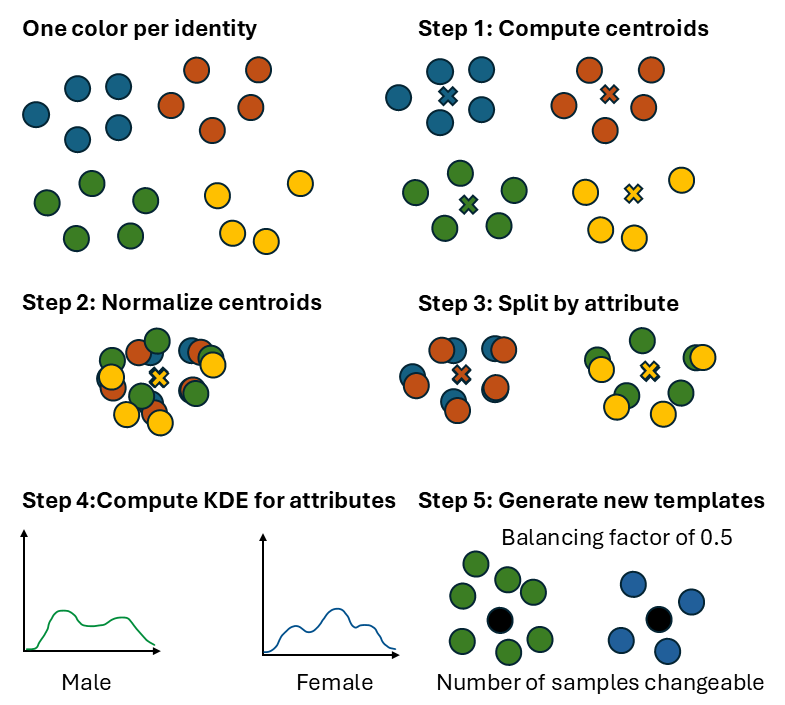}
    \caption{\textbf{Visualization of the template generation process -} (0) Original embeddings with one color per identity,
    (1) Computing centroids for each identity, (2) Normalizing templates by subtracting centroids, (3) Splitting normalized templates by attribute (gender),
    (4) Computing separate Kernel Density Estimation (KDE) models for each attribute, and (5) Generating new templates with a variable balancing factor and controllable sample sizes.}
    \label{fig:kde_process}
\end{figure}

To address the challenge of training a personalized neural network verification model with only one reference sample, we make use of Kernel Density Estimation (KDE) for synthetic template generation, as illustrated in Figure~\ref{fig:kde_process}.
A KDE is trained for each attribute that should be taken into account regarding fairness.
Since in our approach fairness regarding gender is a concern, two KDEs are trained, one for male and one for female.
This approach allows to manipulate the distribution of generated samples. 

As shown in Step 1 of Figure~\ref{fig:kde_process}, for each identity $i$, we compute its
centroid $c_i$ from the set of original embeddings $X_i = \{x_1, \ldots, x_j, \ldots x_n\}$:
\begin{equation}
c_i = \frac{1}{n}\sum_{j} x_j, \quad x_j \in X_i
\end{equation}
The templates are then normalized by subtracting their respective identity centroids (Step 2 in Figure~\ref{fig:kde_process}):
\begin{equation}
\tilde{x}_j = x_j - c_i
\end{equation}
These normalized templates are split by attribute (Step 3 in Figure~\ref{fig:kde_process}), and separate KDE models are trained for each attribute (Step 4),
\begin{equation}
\hat{f}(x) = \frac{1}{nh}\sum_{i} K\left(\frac{x - \tilde{x}_i}{h}\right)
\end{equation}
where $K$ is a Gaussian kernel and $h$ is the bandwidth parameter determined through cross-validation.

%\sh{What kernel K is used? Does the kernel contain parameters? How exactly are the KDE models trained?}

%\sh{How exactly is the bandwidth parameter determined? In particular, what function is optimized in cross validation? The mean integrated squared error? (compare \url{https://en.wikipedia.org/wiki/Kernel_density_estimation\#Bandwidth_selection})}
    
Finally, as depicted in Step 5 of Figure~\ref{fig:kde_process}, new templates $\{x_{sample}\}$ for identity $i$ are generated, based on the corresponding KDE, using a reference template of this identity:
\begin{equation}
x_{i,new} = x_{sample} + c_{i}
\end{equation}
%\sh{What exactly is $x_i$? Why do we not use the centroid $c_i$ here? }
As illustrated in Figure~\ref{fig:kde_process}, this process allows for generating an arbitrary number of synthetic templates
that maintain the identity characteristics while introducing natural variations based on attribute-specific distribution patterns.
    
\subsection{Network Architecture and Training}
With MOTE, we train a dedicated neural network for each enrolled identity to distinguish between genuine and imposter templates. The architecture was
carefully designed to balance complexity, generalization capability, and computational efficiency. We employ a
relatively shallow feed-forward architecture to ensure fast enrollment and inference time.

The network comprises an input layer aligned with the 512-dimensional face embeddings, followed by two hidden layers reducing dimensions to 128 and 64 neurons, respectively, each using ReLU activation and 0.5 dropout, as recommended by Srivastava et al.~\cite{DropoutSrivastavaHKSS14} to mitigate neuron co-adaptation. The final output layer is a single sigmoid-activated neuron yielding a probability that the input template matches the target identity.

%The network architecture consists of an input layer matching the 512-dimensional face embeddings produced by the backbone models.
%We selected this dimension-preserving input to maintain compatibility with existing face recognition system while allowing our approach to be backbone-agnostic. 
%The first hidden layer reduces dimensionality to 128 neurons with ReLU
%activation, followed by a dropout rate of 0.5. The dropout rate of 0.5 was selected following the recommendations of Srivastava et
%al.~\cite{DropoutSrivastavaHKSS14}, who demonstrated its effectiveness for preventing co-adaptation of neurons in intermediate representations.
%The second hidden layer further reduces dimensionality to 64 neurons, also employing ReLU activation and 0.5 dropout. Finally, the output
%layer consists of a single neuron with sigmoid activation, producing a probability score indicating
%whether an input template belongs to the target identity.

We train each identity-specific network using the Adam optimizer~\cite{AdamOptimizerKingmaB14} with default parameters ($\beta_1{=}0.9$, $\beta_2{=}0.999$, $\epsilon{=}10^{-8}$), an initial learning rate of $10^{-2}$, and weight decay of $10^{-3}$. A OneCycleLR scheduler~\cite{PyTorchOneCycleLR} modulates the learning rate following the 1cycle policy~\cite{smithSuperConvergenceVeryFast2018}, enhancing convergence and generalization, particularly for imbalanced binary tasks. The loss function is Binary Cross-Entropy with Logits~\cite{PyTorchBCEWithLogitsLoss}, chosen for its numerical stability and effectiveness in binary classification. Networks are trained for up to 100 epochs with early stopping based on validation loss. Training uses a batch size of 128, with balanced sampling of genuine and imposter templates to mitigate class imbalance.

\subsection{Recognition Workflow}
During authentication, the template extracted from the presented face image is passed through the corresponding identity's trained classifier.
The output probability score determines whether the presented face belongs to the claimed identity:
\begin{equation}
\text{decision} = 
\begin{cases}
\text{genuine}, & \text{if } P(y=1|x) \geq \tau \\
\text{impostor}, & \text{otherwise}
\end{cases}
\end{equation}

where $\tau$ is a decision threshold optimized to balance false matches and false non-matches.

%\sh{How exactly is the decision threshold $\tau$ determined? What exactly is the objective function that is optimized here? }

\section{Experimental Setup}

\subsection{Datasets and Models}
\textbf{Datasets:} To evaluate our MOTE approach, we utilized three widely recognized face recognition
datasets selected for their diverse conditions and demographic distributions. The Adience dataset~\cite{eidingerAgeGenderEstimation2014},
containing 26,580 unconstrained images of 2,284 subjects with gender and age labels, was chosen specifically for its demographic breadth
(gender balance, wide age range) and challenging imaging conditions, including variations in lighting, pose, and quality. Complementing this,
the ColorFeret dataset~\cite{phillipsFERETDatabaseEvaluation1998} provided a more controlled environment with 11,338 images of 994 subjects
under systematic pose, expression, and lighting variations, allowing us to assess performance under constrained conditions and evaluate
pose robustness. Finally, we included the Labeled Faces in the Wild (LFW) dataset~\cite{huangLabeledFacesWild2008}, comprising 13,233
web-collected images of 5,749 subjects (primarily public figures), as it serves as a standard benchmark for face verification, enabling
direct comparison with existing methods~\cite{wangDeepFaceRecognition2021}.

\textbf{Models and Implementation Details:} For face template extraction, we employed two state-of-the-art face recognition models to
generate embeddings. We utilized ArcFace~\cite{dengArcFaceAdditiveAngular2022}, a model known for its discriminative power achieved
through an additive angular margin loss function and its strong performance on standard benchmarks. Alongside this, we used
MagFace~\cite{mengMagFaceUniversalRepresentation2021}, which incorporates magnitude-aware feature embeddings designed to encode both
identity and sample quality information simultaneously; this quality awareness offers a different embedding philosophy and is
particularly relevant for unconstrained datasets like Adience. Both ArcFace and MagFace generate 512-dimensional feature vectors
which serve as input to our identity-specific networks. Importantly, we used the official pre-trained versions of these models without
any fine-tuning on the evaluation datasets. This approach ensures a fair comparison with published results and specifically evaluates
the ability of our MOTE method to enhance established embedding frameworks without requiring their modification.

\subsection{Evaluation Protocol}
\label{sec:EvaluationProtocol}
We established a comprehensive evaluation protocol to assess critical aspects of our MOTE approach, each selected to address specific requirements for modern face recognition systems.
For all experiments, subject-disjoint training and testing sets were used to ensure generalization.

\textbf{Recognition performance}: Following the international ISO standards for biometric performance testing, the verification performance is evaluated in terms of False Match Rate
(FMR) and False Non-Match Rate (FNMR)~\cite{InformationTechnologyBiometric2021}. 
We report FNMR at a fixed FMR of $10^{-3}$ following the recommended guidelines of the European Border Guard Agency FRONTEX  \cite{europeanagencyforthemanagementofoperationalcooperationattheexternalbordersofthememberstatesoftheeuropeanunionGuidelinesProcessingThird2016}.
%We also calculate Area Under Curve (AUC)~\cite{ROCAUC} values to provide a single-value performance metric. 

\textbf{Privacy protection}: 
We assess the privacy protection based on the most successful attack on soft-biometric privacy that we are aware of, inference attacks from Osorio-Roig et al. \cite{osorio-roigAttackFacialSoftBiometric2022}.
These inference attacks are performed on the attribute gender as it is a binary attribute with clear facial differences, making it challenging for soft-biometric privacy-preservation methods.
In simple terms, if an adversary wants to estimate the gender associated with a template with this inference attack, this template is compared against a set of known female and male templates, and the average comparison score per gender is computed. The gender decision is based on the highest average score.

\textbf{Fairness}: We measured demographic fairness using two complementary metrics: False Discrepancy Rate
(FDR)~\cite{defreitaspereiraFairnessBiometricsFigure2022} and 1- Gini Aggregation Rate for Biometric Equitability (1-GARBE)~\cite{howardEvaluatingProposedFairness2022, DBLP:journals/corr/abs-2501-12020}. FDR quantifies the disparity in FMRs and FNMRs across demographic groups (specifically gender), with values closer to 1.0 indicating more equitable performance. GARBE measures inequalities in score distributions using Gini coefficients, with values closer to 0 indicating more equitable treatment. We report 1-GARBE (iGARBE \cite{DBLP:journals/corr/abs-2501-12020}) for consistency with FDR, where high values indicate better fairness.

\textbf{Explainability}: We visualized network attention using Gradient-weighted Class Activation Mapping
(Grad-CAM++)~\cite{chattopadhyayGradCAMImprovedVisual2018} to provide interpretable insights into our model's decision-making process.
This visualization technique highlights regions most influential for classification decisions, addressing the ``black box'' nature of
deep learning models. We selected Grad-CAM++ specifically for its improved localization capabilities compared to standard CAM methods,
as demonstrated by~\cite{selvarajuGradCAMVisualExplanations2020}. However, the choice of explainability methods is flexible in the proposed approach.

\textbf{Resource requirements}: We measured computational efficiency in terms of storage requirements and enrollment time using the Adience, ColorFeret and LFW datasets.
Storage efficiency was quantified as the size in kilobytes (KB) required per identity, providing a direct measure of the system's
scalability for large-scale deployments. Enrollment time was measured in seconds per subject, representing the computational cost
of adding new identities to the system. We employed a ResNet-100 backbone for both the ArcFace and MagFace models. To ensure robust measurements, each enrollment process was repeated 10 times per subject, and the average result across these runs is reported. These practical considerations are essential for assessing real-world deployment feasibility,
particularly for applications with limited computational resources or requiring rapid enrollment processing.
For the experiments, Intel Core i5-12600KF with RTX 2080 Ti and 32gb DDR4 RAM are used.

\section{Results and Discussion}

% \sh{GitHub with code to reproduce results?}

In this section, we present and discuss the results of the experimentation in terms of recognition performance, privacy protection, fairness, explainability, and system efficiency. 

%This section presents the comprehensive evaluation of our proposed model-template (MOTE) based face recognition approach compared to state-of-the-art face recognition systems. We evaluated the approach using three standard face recognition datasets (Adience, ColorFeret, and LFW) across multiple dimensions: recognition performance, privacy protection, fairness, explainability, and system efficiency.

\subsection{Recognition Performance}

\begin{figure*}[!h]
    \centering
    % Column headers
 %   \begin{minipage}[b]{0.2\textwidth}
 %       \centering
 %       \textbf{Adience}
 %   \end{minipage}
 %   \hfill
 %   \begin{minipage}[b]{0.2\textwidth}
 %       \centering
 %       \textbf{ColorFeret}
 %   \end{minipage}
%    \hfill
%    \begin{minipage}[b]{0.25\textwidth}
%        \centering
%        \textbf{LFW}
%    \end{minipage}

    % Add some vertical space between the headers and the images
%    \vspace{0.25em}

%    % First row - ArcFace
%    \vspace{0.25em}
%    \parbox{\textwidth}{\centering \textbf{ArcFace}\\}
%    \rule{\textwidth}{0.5pt} % Horizontal line
%    \vspace{0.25em}
    \begin{subfigure}[b]{0.3\textwidth}
        \vspace{0.2em}
        \centering
        \includegraphics[width=\textwidth]{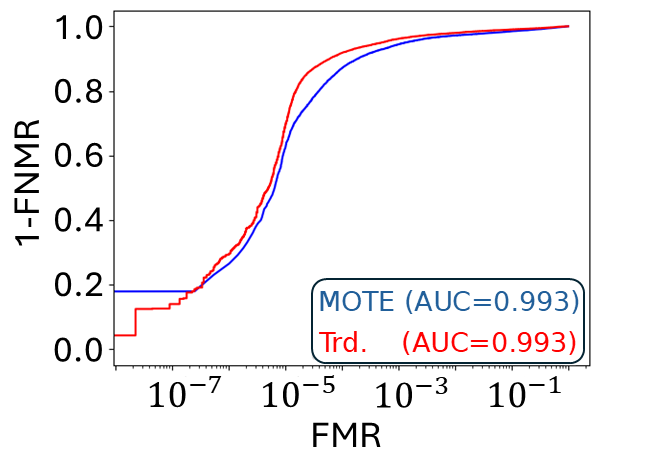}
        \caption{ArcFace on Adience}
        \label{fig:arcface_adience_roc}
    \end{subfigure}
    \hfill
    \begin{subfigure}[b]{0.3\textwidth}
        \vspace{0.2em}
        \centering
        \includegraphics[width=\textwidth]{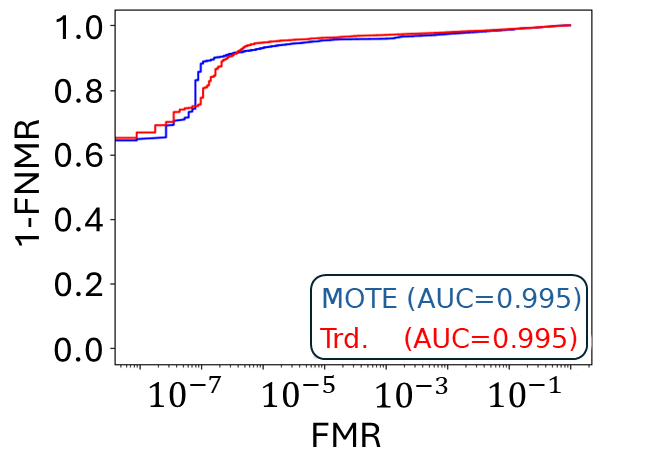}
        \caption{ArcFace on ColorFeret}
        \label{fig:arcface_colorferet_roc}
    \end{subfigure}
    \hfill
    \begin{subfigure}[b]{0.3\textwidth}
        \vspace{0.2em}
        \centering
        \includegraphics[width=\textwidth]{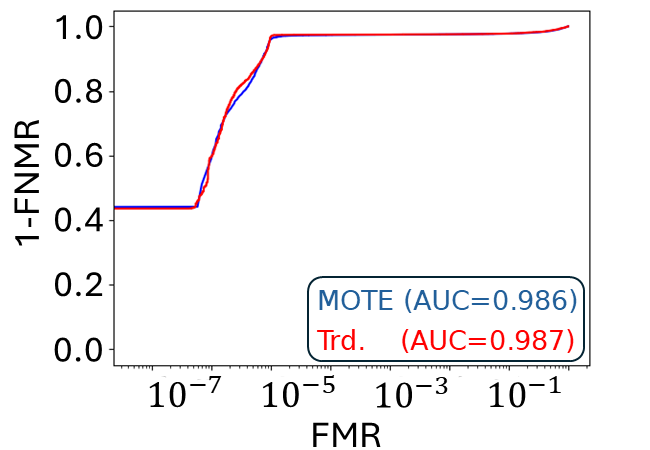}
        \caption{ArcFace on LFW}
        \label{fig:arcface_lfw_roc}
    \end{subfigure}

    % Second row - MagFace
%    \vspace{0.5em}
%    \parbox{\textwidth}{\centering \textbf{MagFace}\\}
%    \rule{\textwidth}{0.5pt} % Horizontal line
    \vspace{0.25em}
    \begin{subfigure}[b]{0.3\textwidth}
        \vspace{0.2em}
        \centering
        \includegraphics[width=\textwidth]{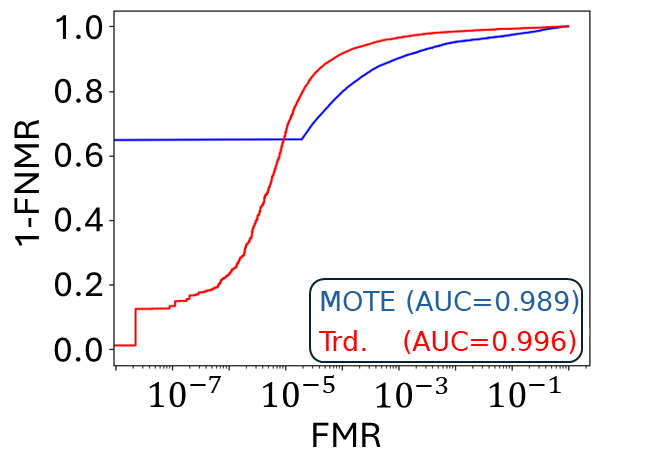}
        \caption{MagFace on Adience}
        \label{fig:magface_adience_roc}
    \end{subfigure}
    \hfill
    \begin{subfigure}[b]{0.3\textwidth}
        \vspace{0.2em}
        \centering
        \includegraphics[width=\textwidth]{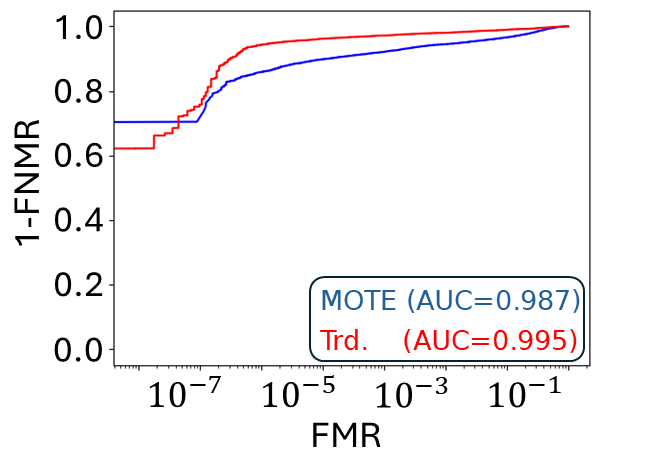}
        \caption{MagFace on ColorFeret}
        \label{fig:magface_colorferet_roc}
    \end{subfigure}
    \hfill
    \begin{subfigure}[b]{0.3\textwidth}
        \vspace{0.2em}
        \centering
        \includegraphics[width=\textwidth]{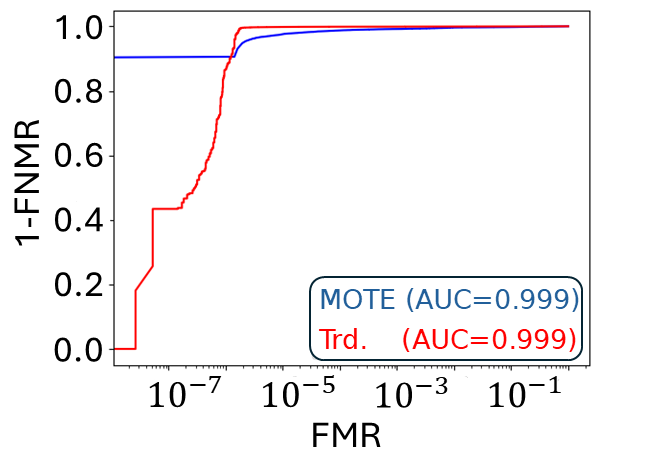}
        \caption{MagFace on LFW}
        \label{fig:magface_lfw_roc}
    \end{subfigure}

    \caption{\textbf{Analysing Verification Performance in Terms Of ROC Curves -} The FNMR is reported over various FMR, comparing the proposed MOTE approach against traditional (Trd.) face recognition methods (either ArcFace or MagFace) across multiple datasets. Over all dataset and face recognition systems, a weaker but comparable performance is observable.}
    \label{fig:comparison_roc}
\end{figure*}

To show that the performance of MOTE is comparable to state-of-the-art face recognition systems, Table \ref{tab:accuracy} presents the accuracy comparison between our approach and existing systems and Table \ref{tab:fnmr} presentes the verification performance in terms of FNMR at $10^{-3}$ FMR.
For a more comprehensive performance evaluation, the Receiver Operating Characteristic (ROC) curves and their corresponding Area Under the Curve (AUC) values are shown in Figure~\ref{fig:comparison_roc}.
In all cases of combinations of face recognition systems and datasets, the traditional systems perform slightly better than then MOTE variants. This is to be expected as the FRS is carefully optimized for the recognition task while the MOTE approach uses one training procedure for all models. However, MOTE still leads to comparable performances. 
%Table~\ref{tab:fnmr} reveals that the performance gap between male and female subjects is reduced more with MOTE in 4 out of 6 experiments.
%%% OLD.
% In Table \ref{tab:fnmr}, the effect of this slight performance is already observable as the male and female performance equalizes more significantly.
While the performance of MOTE drops slightly compared to the traditional works, it leads to a  fairer performance as it will be further analysed in the Section \ref{sec:Fairness}.

%we analyzed the Receiver Operating Characteristic (ROC) curves and their corresponding Area Under the Curve %(AUC) values. As shown in Figure~\ref{fig:comparison_roc}, the ROC curves demonstrate remarkably similar behavior
%between our proposed method and the baseline models across all datasets, indicating comparable discrimination capabilities.

%The model-template based approach demonstrates competitive performance when compared to state-of-the-art face recognition systems (ArcFace and MagFace) across all tested datasets. Table \ref{tab:accuracy} presents the accuracy comparison between our approach and existing systems.

\begin{table}[ht]
    \renewcommand{\arraystretch}{1.1}
    \centering
    %\normalsize
    \caption{\textbf{Analysis of Recognition Accurracy -} The recognition accuracy between our MOTE approach and the traditional
    face recognition methods are compared across multiple datasets. MOTE performs similarly to the traditional models.}
    \label{tab:accuracy}
    \begin{tabular}{llrr}
    \Xhline{2\arrayrulewidth}
    \textbf{Dataset} & \textbf{Method} & \textbf{MOTE} & \textbf{Traditional} \\ \hline
    \multirow{2}{*}{Adience} & ArcFace & 99.956\% & 99.991\% \\
                             & MagFace & 99.462\% & 99.989\% \\ \hline
    \multirow{2}{*}{ColorFeret} & ArcFace & 99.992\% & 99.990\% \\
                                & MagFace & 99.761\% & 99.990\% \\ \hline
    \multirow{2}{*}{LFW} & ArcFace & 99.998\% & 99.998\% \\
                         & MagFace & 99.424\% & 100.000\% \\ 
     \Xhline{2\arrayrulewidth}
    \end{tabular}
    
\end{table}

%\captionsetup{font=scriptsize}
%\captionsetup[subfigure]{labelformat=empty}

%When compared with ArcFace, our approach achieved highly competitive AUC values: 0.99269 vs. 0.99542 on Adience, 0.99481 vs. 0.99496 on ColorFeret, and 0.98635 vs. 0.98716 on LFW. While the performance gap with MagFace was slightly larger, our approach still achieved strong results with AUC values of 0.98905 vs. 0.99663 on Adience, 0.98722 vs. 0.99535 on ColorFeret, and 0.99891 vs. 0.99984 on LFW.

%We also examined the False Non-Match Rates (FNMR) at a fixed False Match Rate (FMR) of $10^{-3}$, a threshold recommended by Frontex for automated border control
%systems~\cite{europeanagencyforthemanagementofoperationalcooperationattheexternalbordersofthememberstatesoftheeuropeanunionGuidelinesProcessingThird2016}.
%Our approach showed slightly higher but still competitive FNMR values compared to traditional systems, as detailed in Table \ref{tab:fnmr}.

\begin{table}[ht]
    \setlength{\tabcolsep}{2pt}
    \renewcommand{\arraystretch}{1.1}
    \centering
    \normalsize
    \caption{\textbf{Analyis of Verification Performance - } The verification performance between MOTE and traditional methods across different face recognition models,
    datasets, and demographic groups are investigated. Values represent FNMRs at a fixed FMR of $10^{-3}$. MOTEs performance lacks a little but becomes more fair.}
    \label{tab:fnmr}
    \begin{tabular}{llcccc}
    \Xhline{2\arrayrulewidth}
    \textbf{Dataset} & \textbf{Group} & \multicolumn{2}{c}{\textbf{ArcFace}} & \multicolumn{2}{c}{\textbf{MagFace}} \\
    \cmidrule(rl){3-4} \cmidrule(rl){5-6}
     & & MOTE & Traditional & MOTE & Traditional \\ \hline
    \multirow{3}{*}{Adience} & All & 0.0558 & 0.0387 & 0.0987 & 0.0355 \\
     & Male & 0.0566 & 0.0340 & 0.0967 & 0.0336 \\
     & Female & 0.0566 & 0.0424 & 0.0976 & 0.0364 \\ \hline
    \multirow{3}{*}{ColorFeret} & All & 0.0418 & 0.0299 & 0.0794 & 0.0285 \\
     & Male & 0.0424 & 0.0299 & 0.0812 & 0.0279 \\
     & Female & 0.0398 & 0.0309 & 0.0722 & 0.0303 \\ \hline
    \multirow{3}{*}{LFW} & All & 0.0260 & 0.0256 & 0.0085 & 0.0011 \\
     & Male & 0.0259 & 0.0258 & 0.0077 & 0.0013 \\
     & Female & 0.0260 & 0.0243 & 0.0096 & 0.0003 \\ 
     \Xhline{2\arrayrulewidth}
    \end{tabular}
    
\end{table}

\subsection{Privacy Protection}

A key advantage of the proposed MOTE approach is its enhanced privacy protection against soft-biometric attribute extraction attacks. The system's resistance to gender inference attacks is evaluated by simulating attempts to extract gender information from facial templates and model decisions as explained in Section \ref{sec:EvaluationProtocol}.

Table \ref{tab:privacy} compares the vulnerability to gender inference attacks across different approaches. The results show that while both the traditional models and existing privacy-enhancing techniques (PFRNet \cite{bortolatoLearningPrivacyenhancingFace2020} and PE-MIU \cite{terhorstPEMIUTrainingFreePrivacyEnhancing2020}) remain highly vulnerable to gender inference attacks, our proposed method successfully mitigates this vulnerability by reducing attack success rates to near-random levels (balanced accuracies consistently around 50\%).

\begin{table}[ht]
    \setlength{\tabcolsep}{3pt}
    \renewcommand{\arraystretch}{1.1}
    \centering
    \normalsize
    \caption{\textbf{Analysis of Privacy-Protection - } The vulnerability to gender inference attacks across different face recognition methods and datasets is investigated. Values represent attack success rates (\%), where 50\% indicates ideal privacy protection. MOTE methods consistently demonstrate near-optimal privacy protection.}
    \begin{tabular}{lcccc}
    \hline
    \textbf{Dataset} & \textbf{MOTE} & \textbf{MOTE} & \textbf{PFRNet} & \textbf{PE-MIU} \\
     & ArcFace & MagFace & \cite{bortolatoLearningPrivacyenhancingFace2020} & \cite{terhorstPEMIUTrainingFreePrivacyEnhancing2020} \\
    \hline
    Adience & \textbf{50.29} & \textbf{50.10} & 79.60 & 88.91 \\
    ColorFeret & \textbf{50.02} & \textbf{50.00} & -- & 93.34 \\
    LFW & \textbf{50.00} & \textbf{50.00} & 83.67 & 90.16 \\ \hline
    \end{tabular}
    \label{tab:privacy}
\end{table}

Since MOTE does not store vector templates in the database, it is not vulnerable to soft-biometric privacy attacks \cite{terhorstUnsupervisedPrivacyenhancementFace2019, terhorstSuppressingGenderAge2019, bortolatoLearningPrivacyenhancingFace2020} that predict attributes based on vector representations.
This leaves a potential vulnerability to inference-based attacks \cite{osorio-roigAttackFacialSoftBiometric2022}.
However, our analysis revealed that MOTE achieves robust privacy protection by creating a decision boundary that effectively neutralizes gender inference attacks, reducing them to random guessing performance as personalized models are not calibrated amoung each other. Unlike traditional systems, where templates inadvertently encode soft-biometric attributes, our personalized classifiers successfully obfuscate gender information while maintaining high recognition performance.

This privacy protection stems from a fundamental difference in output characteristics: when attacking MOTE on a given dataset, all gender predictions collapse to a single class with unalgined comparison scores,
effectively destroying any meaningful inference capability. In contrast, traditional systems produce predictions
whose gender distribution closely mirrors the actual distribution in the attack dataset, indicating a persistent gender
information leak in their encoded templates.

\begin{table}[t]
    \setlength{\tabcolsep}{3pt}
    \renewcommand{\arraystretch}{1.1}
    \centering
    \normalsize
    \caption{\textbf{Analysis of Fairness -} The fairness regarding gender is measured in FDR and iGARBE is across different balancing factors during enrolment training. Higher values indicate better fairness. Bold values represent equal or higher performance of MOTE over the traditional models (Trd). MOTE shows higher fairness for various scenarios, even without considering that the fairness can be tuned for each individual separately.}
    \label{tab:gender_fairness}
        \begin{tabular}{lllcccc}
            \hline
            \textbf{Model} & \textbf{Data} & \textbf{Balancing} & \multicolumn{2}{c}{\textbf{FDR}} & \multicolumn{2}{c}{\textbf{iGARBE}} \\
            \cmidrule(rl){4-5} \cmidrule(rl){6-7}
            & & \textbf{Factor} & MOTE & Trd & MOTE & Trd \\
            \hline
            \multirow{15}{*}{\rotatebox{90}{ArcFace}} & \multirow{5}{*}{\rotatebox{90}{Adience}} 
            & 0.0 & \textbf{0.998} & \multirow{5}{*}{0.996} & \textbf{0.979} & \multirow{5}{*}{0.945} \\
            & & 0.4 & \textbf{0.997} & & \textbf{0.972} & \\
            & & 0.5 & \textbf{1.000} & & \textbf{0.999} & \\
            & & 0.6 & \textbf{0.998} & & \textbf{0.982} & \\
            & & 1.0 & \textbf{0.998} & & \textbf{0.984} & \\
            \cline{2-7}
            & \multirow{5}{*}{\rotatebox{90}{ColorFeret}}
            & 0.0 & \textbf{0.999} & \multirow{5}{*}{0.999} & 0.986 & \multirow{5}{*}{0.990} \\
            & & 0.4 & \textbf{0.999} & & 0.977 & \\
            & & 0.5 & \textbf{0.999} & & 0.947 & \\
            & & 0.6 & \textbf{0.999} & & 0.951 & \\
            & & 1.0 & \textbf{0.999} & & 0.978 & \\
            \cline{2-7}
            & \multirow{5}{*}{\rotatebox{90}{LFW}}
            & 0.0 & \textbf{1.000} & \multirow{5}{*}{0.999} & \textbf{0.994} & \multirow{5}{*}{0.985} \\
            & & 0.4 & \textbf{1.000} & & \textbf{0.990} & \\
            & & 0.5 & \textbf{1.000} & & \textbf{0.988} & \\
            & & 0.6 & \textbf{1.000} & & \textbf{0.999} & \\
            & & 1.0 & \textbf{0.999} & & \textbf{0.986} & \\
            \hline
            \multirow{15}{*}{\rotatebox{90}{MagFace}} & \multirow{5}{*}{\rotatebox{90}{Adience}}
            & 0.0 & \textbf{1.000} & \multirow{5}{*}{0.999} & \textbf{0.999} & \multirow{5}{*}{0.980} \\
            & & 0.4 & 0.997 & & \textbf{0.986} & \\
            & & 0.5 & \textbf{0.999} & & \textbf{0.998} & \\
            & & 0.6 & \textbf{1.000} & & \textbf{0.999} & \\
            & & 1.0 & \textbf{0.999} & & \textbf{0.997} & \\
            \cline{2-7}
            & \multirow{5}{*}{\rotatebox{90}{ColorFeret}}
            & 0.0 & 0.998 & \multirow{5}{*}{0.999} & \textbf{0.987} & \multirow{5}{*}{0.970} \\
            & & 0.4 & 0.996 & & \textbf{0.970} & \\
            & & 0.5 & 0.995 & & 0.967 & \\
            & & 0.6 & 0.998 & & \textbf{0.981} & \\
            & & 1.0 & 0.997 & & \textbf{0.973} & \\
            \cline{2-7}
            & \multirow{5}{*}{\rotatebox{90}{LFW}}
            & 0.0 & \textbf{1.000} & \multirow{5}{*}{0.999} & \textbf{0.997} & \multirow{5}{*}{0.692} \\
            & & 0.4 & \textbf{1.000} & & \textbf{0.993} & \\
            & & 0.5 & \textbf{0.999} & & \textbf{0.946} & \\
            & & 0.6 & \textbf{0.999} & & \textbf{0.918} & \\
            & & 1.0 & 0.998 & & \textbf{0.899} & \\
            \hline
        \end{tabular}
\end{table}

\subsection{Fairness Evaluation}
\label{sec:Fairness}

The fairness of the proposed MOTE approach in comparison to two traditional approachs, ArcFace and MagFace, is analysed and shown in Table \ref{tab:gender_fairness} with respect to the FDR and iGARBE fairness measures on three datasets.
We further analyzed how gender distribution in training data affects the system's fairness by systematically varying the ratio of male to female
embeddings in the training set. We selected specific balancing factors (0.0, 0.4, 0.5, 0.6, and 1.0) to comprehensively investigate fairness
implications across different gender distributions. A balancing factor of 0.0 has the consequence that only female templates are sampled, while a balancing factor of 1.0 has the consequence of only sampling male templates. These ratios were strategically chosen to examine: (1) extreme cases with
single-gender training (0.0 and 1.0), which reveal potential worst-case biases; (2) balanced distribution (0.5), representing the
theoretical ideal for unbiased learning; and (3) slightly imbalanced distributions (0.4 and 0.6), which better reflect real-world
demographic variations in many datasets and application contexts. This range allows us to assess whether achieving optimal fairness
requires perfect balance or if systems can tolerate moderate imbalance. %Table~\ref{tab:gender_fairness} presents gender fairness

\begin{figure}
    \centering
    \includegraphics[width=0.5\textwidth]{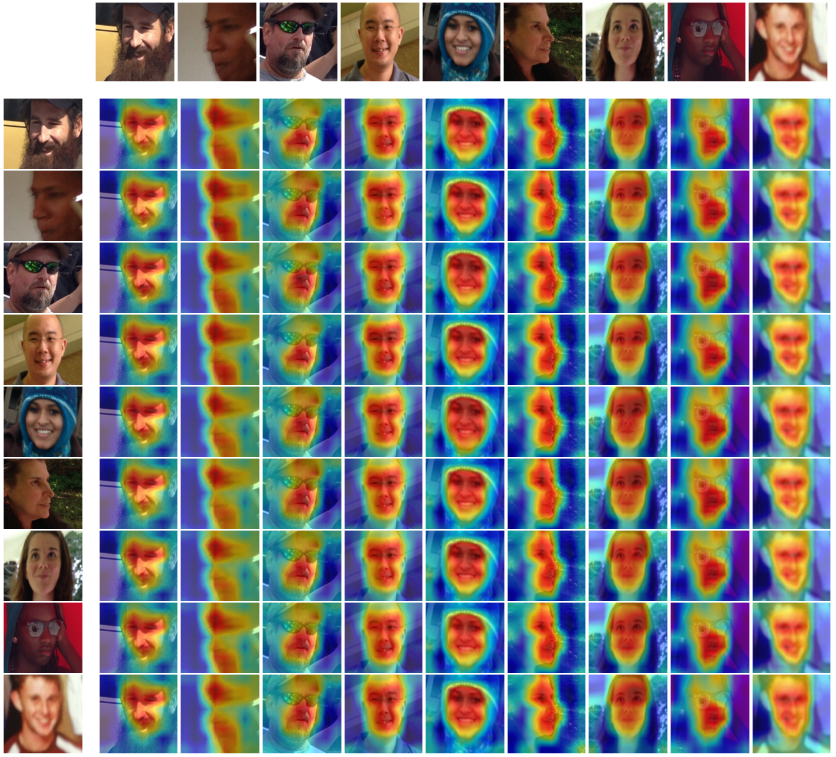}
    \caption{\textbf{Analysis of Explainability -} Grad-CAM++ Visualization of MOTE demonstrating cross-subject behavior. The leftmost column shows template images used to train individual classifiers, featuring varying lighting conditions, backgrounds, and poses. The top row displays the corresponding test images. The heatmaps in subsequent rows demonstrate how each personlized model attends to facial features when making decisions, with red regions indicating areas of highest importance for the identity verification.}
    \label{fig:gradcam_paper_result}
\end{figure}

\begin{table*}[tb]
    \setlength{\tabcolsep}{4pt}
    \renewcommand{\arraystretch}{1.1}
    \centering
    \normalsize
    \caption{\textbf{Anaylsis of System Efficiency -} Storage requirements and enrollment times are reported. For the traditional ArcFace and MagFace approaches, the same architecture is used, leading to similar enrollment times.}
    \label{tab:system_efficiency}
    \begin{tabular}{lcccc}
    \Xhline{2\arrayrulewidth}
        \textbf{Method} & \textbf{Storage per Identity} (in KB) & \multicolumn{3}{c}{\textbf{Enrollment Time} (in sec)} \\
        \cmidrule(rl){3-5}
        &  & Adience & ColorFeret & LFW \\
        \hline
        Traditional Template (ArcFace) & 39.2 KB & 0.121 $\pm$ 0.034 & 0.096 $\pm$ 0.027 & 0.147 $\pm$ 0.044 \\
        \multirow{1}{*}{MOTE (ArcFace)} & \multirow{1}{*}{297 KB} & 4.282 $\pm$ 0.855 & 5.952 $\pm$ 0.845 & 6.632 $\pm$ 3.871 \\
        Traditional Template (MagFace) & 39.2 KB & 0.138 $\pm$ 0.039 & 0.101 $\pm$ 0.036 & 0.165 $\pm$ 0.051 \\
        \multirow{1}{*}{MOTE (MagFace)} & \multirow{1}{*}{297 KB} & 4.452 $\pm$ 1.493 & 5.082 $\pm$ 1.024 & 5.032 $\pm$ 1.713 \\
        \Xhline{2\arrayrulewidth}
    \end{tabular}
\end{table*}

Our results demonstrate that the proposed MOTE approach consistently outperforms the traditional methods in terms of gender fairness as measured
by both FDR and iGABRE metrics. The relationship between gender
distribution in training data and fairness outcomes varies by dataset:

For the Adience dataset, where image quality is relatively evenly distributed between genders, balanced gender distributions
(balancing factor of 0.5-0.6) often yield the best fairness scores, particularly for iGARBE metrics where we observe peaks at these ratios.

In contrast, for the LFW dataset, where female subjects exhibit greater variability due to factors such as makeup and hairstyle, we observe a
different pattern. FDR scores remain high across all balancing factors, but iGARBE metrics show a gradual decline as the balancing factor increases from
0.0 to 1.0, with the highest scores at a lower balancing factor. This suggests that including more female samples in training data helps the model
better account for the greater variability in female facial features. When these additional variability factors aren't adequately represented
in training (as the balancing factor increases), fairness performance decreases accordingly.

The ColorFeret dataset shows more complex patterns, as the high variability of head poses highlights the major challenge.
Thus, the gender sampling is strongly dependent on these randomly chosen head poses, resulting in some metrics peaking at extreme ratios (0.0 or 1.0) rather than balanced ratios.

Overall, our MOTE approach demonstrates robust gender fairness across multiple datasets and baseline models, with particularly notable improvements over traditional methods when evaluated on the LFW dataset using the MagFace model, where we observed up to
30.5\% improvement in iGARBE (0.997 vs. 0.692 at 0.0 of balancing factor).

Overall, our findings indicate that the optimal gender distribution in training data varies by dataset, model, and evaluation metric, reflecting the complex interaction between data characteristics and fairness outcomes. However, the MOTE approach demonstrates robust gender fairness across these variations.

\subsection{Explainability Analysis}

To understand the decision-making process of our model-template based approach, we employed Gradient-weighted Class Activation Mapping (Grad-CAM++) visualization. Figure \ref{fig:gradcam_paper_result} shows the attention heatmaps generated from our models verification decisions.

The visualization reveals that our approach consistently focuses on crucial facial regions, particularly the central facial area encompassing the eyes, nose, and mouth. The MOTE models demonstrate remarkable adaptability in their attention mechanisms, dynamically adjusting their focus when confronted with occlusions or varying pose angles. This natural alignment with human perceptual understanding of face recognition suggests that our models have learned to meaningfully represent the identity.

% \sh{So what? Why does SMOTE produce better explanations than previous approaches? The explanations look ``boring''/uninformative. Each explanation in a column looks (essentially) the same. Of course, a face recognition system will consider faces. But how does the classifier distinguish or not distinguish male/female? How does the classifier distinguish identities? }

\subsection{System Efficiency}

Finally, the practical aspects of our approach are evaluated: storage requirements and enrollment speed.
Table \ref{tab:system_efficiency} provides a comprehensive comparison of these system efficiency metrics across different face recognition systems.

%\begin{table*}[htb]
%    \centering
%    \caption{System Efficiency: Storage Requirements and enrollment Time. (Traditional - same %architecture and thus same numbers), explain SDV}
%    \begin{adjustbox}{width=0.9\textwidth}
%    \begin{tabular}{lccccccc}
%    \setlength{\tabcolsep}{3pt}
%        \renewcommand{\arraystretch}{1.1}
%        \multirow[b]{3}{*}{\textbf{Method}} & \multirow{3}{*}{\textbf{\begin{tabular}[c]{@{}c@{}}\\Storage per\\Identity (KB)\end{tabular}}} & \multicolumn{6}{c}{\textbf{Enrollment Time (seconds)}} \\
%        \cline{3-8}
%        &  & \multicolumn{2}{c}{\textbf{Adience}} & \multicolumn{2}{c}{\textbf{ColorFeret}} & \multicolumn{2}{c}{\textbf{LFW}} \\
%        &  & Avg & & Avg & Std Dev & Avg & Std Dev \\
%        \hline
%        Original Template (ArcFace) & 39.2 KB & 0.452 & 0.094 & 0.452 & 0.094 & 0.452 & 0.094 \\
%        Original Template (MagFace) & 39.2 KB & 0.452 & 0.094 & 0.452 & 0.094 & 0.452 & 0.094 \\
%        \hline
%        \multirow{1}{*}{MOTE (ArcFace)} & \multirow{1}{*}{297 KB} & 4.282 & 0.855 & 5.952 & 0.845 & 6.632 & 3.871 \\
%        \multirow{1}{*}{MOTE (MagFace)} & \multirow{1}{*}{297 KB} & 4.452 & 1.493 & 5.082 & 1.024 & 5.032 & 1.713 \\
%        \hline
%    \end{tabular}
%    \end{adjustbox}
%    \label{tab:system_efficiency}
%\end{table*}

As shown in the table, the proposed MOTE approach requires approximately 7.6 times more storage per identity (297 KB versus 39.2 KB for traditional template methods), though these requirements remain manageable for modern systems.
For enrollment speed, MOTE adds a one-time (small-scale) classifier training cost to the traditional template extraction process,
which varies depending on both the template extractor used and the dataset.
The overall training times range from 3.83 to 6.18 seconds with ArcFace, and 4.00 to 4.63 seconds with MagFace across the different datasets resulting from the used hardware (see Section \ref{sec:EvaluationProtocol}).
While this represents a slower enrollment process compared to traditional template extraction using a ResNet100 backbone (which takes approximately 130 ms),
it is a one-time cost that does not affect subsequent recognition operations. 
For inference speed, per‐embedding decision time remains almost constant at approximately 0.31 ms across all experiment scenarios, ensuring real‐time performance.

\section{Conclusion}

In this work, we proposed MOTE (Model-Template), a novel approach to face recognition that replaces traditional fixed vector templates with small, individualized neural networks. By doing so, MOTE addresses several pressing challenges in face recognition, including concerns surrounding privacy, fairness, and explainability, while maintaining recognition accuracy comparable to traditional recognition systems such as ArcFace and MagFace.
Our extensive experiments show that MOTE significantly improves privacy protection by reducing  inference attack success rates from over 85\% to near-random chance. Furthermore, the system either preserves or enhances fairness metrics, and uniquely enables fine-grained fairness adjustments at the level of individual identities during the enrollment process. The structure of MOTE also facilitates the application of established explainability methods, such as Grad-CAM++, allowing for clearer interpretation of the decision-making process behind recognition outcomes.
While MOTE introduces additional resource requirements, including approximately 7.5 times more storage per identity and increased enrollment time, these costs are acceptable within the context of small- and medium-scale applications where privacy and fairness are critical requirements. Overall, MOTE represents a promising step toward more responsible face recognition technologies, offering a practical solution for settings where ethical considerations are as important as technical performance.

%------------------------------------------------------------------------
%\section{Only for Final copy}

%You must include your signed IEEE copyright release form when you submit
%your finished paper. We MUST have this form before your paper can be
%published in the proceedings.

\clearpage
\printbibliography

%{\small
%\bibliographystyle{ieee}
%\bibliography{egbib}
%}

\end{document}